\def\year{2021}\relax
\documentclass[letterpaper]{article} 
\usepackage{aaai21}  
\usepackage{times}  
\usepackage{helvet} 
\usepackage{courier}  
\usepackage[hyphens]{url}  
\usepackage{graphicx} 
\usepackage{caption}
\usepackage{subfigure}
\usepackage{natbib}  
\usepackage{caption} 
\usepackage{amsmath}
\usepackage{booktabs}
\usepackage{algorithm}
\usepackage{algorithmic}
\usepackage{subfigure}
\usepackage{xspace}
\usepackage{amssymb}
\usepackage{enumitem}
\usepackage{multirow}
\usepackage{siunitx}
\usepackage{bm}
\usepackage{dsfont}
\newcommand{\sys}{{MFES-HB}\xspace}

\title{MFES-HB: Efficient Hyperband with Multi-Fidelity Quality Measurements}

\author{
    Yang Li,\textsuperscript{\rm 1,4}
    Yu Shen,\textsuperscript{\rm 1,4}
    Jiawei Jiang,\textsuperscript{\rm 2} Jinyang Gao,\textsuperscript{\rm 5} 
    Ce Zhang,\textsuperscript{\rm 2} Bin Cui\textsuperscript{\rm 1,3}
    \\
}
\affiliations{
    \textsuperscript{\rm 1}Key Laboratory of High Confidence Software Technologies (MOE), School of EECS, Peking University, Beijing, China\\
    \textsuperscript{\rm 2}Department of Computer Science, Systems Group, ETH Zurich, Switzerland\\ 
    \textsuperscript{\rm 3}Institute of Computational Social Science, Peking University (Qingdao), China \\
    \textsuperscript{\rm 4}AI Platform, Kuaishou Technology, Beijing, China \\
    \textsuperscript{\rm 5}Alibaba Group, Hangzhou, China\\
    
    \{liyang.cs,shenyu,bin.cui\}@pku.edu.cn,   
    \{jiawei.jiang, ce.zhang\}@inf.ethz.ch,
    jinyang.gjy@alibaba-inc.com
}

\begin{document}

\maketitle

\begin{abstract}
Hyperparameter optimization (HPO) is a fundamental problem in automatic machine learning (AutoML).
However, due to the expensive evaluation cost of models (e.g., training deep learning models or training models on large datasets), vanilla Bayesian optimization (BO) is typically computationally infeasible. 
To alleviate this issue, Hyperband (HB) utilizes the early stopping mechanism to speed up configuration evaluations by terminating those badly-performing configurations in advance.
This leads to two kinds of {\em quality measurements:} 
(1) many low-fidelity measurements for configurations that get early-stopped, and (2) few high-fidelity measurements for configurations that are evaluated without being early stopped.
The state-of-the-art HB-style method, BOHB, aims to combine the benefits of both BO and HB. 
Instead of sampling configurations randomly in HB, BOHB samples configurations based on a BO surrogate model, which is constructed with the high-fidelity measurements only.
However, the scarcity of high-fidelity measurements greatly hampers the efficiency of BO to guide the configuration search.

In this paper, we present MFES-HB, an efficient Hyperband method that is capable of utilizing {\em both} the high-fidelity and low-fidelity measurements to accelerate the convergence of HPO tasks.
Designing MFES-HB is not trivial as the low-fidelity measurements can be biased yet informative to guide the configuration search.
Thus we propose to build a Multi-Fidelity Ensemble Surrogate (MFES) based on the generalized Product of Experts framework, which can integrate useful information from multi-fidelity measurements effectively.
The empirical studies on the real-world AutoML tasks demonstrate that \sys can achieve $3.3-8.9\times$ speedups over the state-of-the-art approach --- BOHB.

\end{abstract}

\section{Introduction}
The performance of Machine Learning (ML) models heavily
depends on the specific choice of hyperparameter configurations.
As a result, automatically tuning the hyperparameters has attracted lots of interest from both academia and industry, and has become an indispensable component in modern AutoML systems~\cite{automl_book,automl,DBLP:journals/corr/abs-1904-12054}.
Hyperparameter optimization (HPO) is often a computationally-intensive process as one often needs to try hyperparameter configurations iteratively, and evaluate each configuration by training and validating the corresponding ML model.
However, for ML models that are computationally expensive to train (e.g., deep learning models or models trained on large datasets), vanilla Bayesian optimization (BO)~\cite{hutter2011sequential,bergstra2011algorithms,snoek2012practical}, one of the most prevailing frameworks in solving the HPO problem, suffers from the low-efficiency issue due to insufficient configuration evaluations within a limited budget. 

Instead, Hyperband (HB)~\cite{li2018hyperband} is a popular alternative, which uses the early stopping strategy to speed up configuration evaluation.
In HB, the system dynamically allocates resources to a set of configurations drawn from a uniform distribution, and uses successive halving~\cite{jamieson2016non} to early stop the poorly-performing configurations after measuring 
their quality periodically.
Among many efforts to improve Hyperband~\cite{klein2016learning,pmlr-v80-falkner18a}, BOHB~\cite{pmlr-v80-falkner18a} combines the benefits from both Hyperband and traditional BO. It replaces the configuration sampling procedure in HB from the uniform distribution to a BO surrogate model ---\emph{TPE}~\cite{bergstra2011algorithms}, which is fitted on those quality measurements obtained from the evaluations without being early stopped.

Due to the successive halving strategy of HB, most configuration evaluations end up being early stopped by the system, thus creating two kinds of \emph{quality measurements}: (1) many low-fidelity quality measurements of these early-stopped configurations, and (2) few high-fidelity quality measurements of configurations that are evaluated with complete training resources.
One fundamental limitation of BOHB lies in the fact that the BO component only utilizes the few high-fidelity measurements to sample configurations, which are insufficient to train a BO surrogate that models the objective function in HPO well. 
Consequently, like vanilla BO, sampling configurations in BOHB also suffers from the low-efficiency issue.
{\em Can we also take advantage of the low-fidelity quality measurements,
which are ignored by the existing methods, to further speed up the Hyperband-style methods?}
In this paper, our goal is to investigate a new Hyperband-style method, which is capable of utilizing \emph{the multi-fidelity quality measurements}: both the high-fidelity measurements and the low-fidelity measurements, to accelerate the HPO process.

{\bf (Opportunities and Challenges)} Taking advantage of the multi-fidelity quality measurements poses unique opportunities and challenges.
Intuitively, numerous low-fidelity measurements are obtained from the early-stopped evaluations, which can boost the total number of measurements that BO can use. 
The low-fidelity measurements obtained with partial training resources yield a biased surrogate of the high-fidelity quality measurements. 
Nevertheless, due to the relevancy between early-stopped evaluations and complete evaluations, they can still reveal some useful information about the objective function. 
Therefore, there is great potential for utilizing the low-fidelity measurements to accelerate HPO.
However, {\em if we cannot balance the benefits and biases from the low-fidelity measurements well}, we might be misled by the harmful information towards a wrong objective function. 

In this paper, we propose MFES-HB, an efficient Hyperband method, which is capable of utilizing the planetary unexploited multi-fidelity measurements to significantly accelerate the convergence of configuration search.
To utilize the multi-fidelity measurements without introducing the biases from the low-fidelity measurements, we first train multiple base surrogates on these measurements grouped by their fidelities. 
Then we propose the {\em Multi-Fidelity Ensemble Surrogate (MFES)} that is used in the BO framework to sample configurations. 
Concretely, to make the best usage of those biased yet informative low-fidelity measurements, MFES uses the generalized Product of Experts (gPoE)~\cite{cao2014generalized} to combine these base surrogates, and the contribution of each base surrogate to MFES can be adjusted based on their performance when approximating the objective function. 
Therefore, the heterogeneous information among multi-fidelity measurements could be automatically extracted by MFES in a reliable yet efficient way.
The empirical studies on the real-world HPO tasks demonstrate the superiority of the proposed method over competitive baselines.
\sys can achieve an order of magnitude speedups compared with Hyperband, and $3.3-8.9\times$ speedups over the state-of-the-art method --- BOHB.


\section{Related Work}
\label{sec2}

\textbf{Bayesian Optimization (BO).}
Machine learning (ML) has made great strides in many application areas, e.g., recommendation, computer vision, etc~\cite{goodfellow2016deep,he2017neural,jiang2017tencentboost,ma2019mmm,Wu2020,DBLP:journals/chinaf/ZhangJSC20}.
BO has been successfully applied to tune the hyperparameters of ML models. 
The main idea of BO is to use a probabilistic surrogate model $M:p_{M}(f|\bm{x})$ to describe the relationship between a hyperparameter configuration $\bm{x}$ and its performance $f(\bm{x})$ (e.g., validation error), and then utilizes this surrogate to guide the configuration search (See more details about BO in Section 3).
Spearmint~\cite{snoek2012practical} uses Gaussian process (GP) to model $p_{M}(f|\bm{x})$ as a Gaussian distribution, and TPE \cite{bergstra2011algorithms} employs a tree-structured Parzen density estimators to model $p_{M}(f|\bm{x})$.
Lastly, SMAC \cite{hutter2011sequential} adopts a modified random forest to yield an uncertain estimate of $f(\bm{x})$.
An empirical evaluation of the three methods~\cite{eggensperger2013towards} shows that SMAC performs the best on the benchmarks with high-dimensional and complex hyperparameter space that includes categorical and conditional hyperparameters, closely followed by TPE. 
Spearmint only works well with low-dimensional continuous hyperparameters, and cannot apply to complex configuration space easily.

\noindent
\textbf{Early stopping mechanism} that stops the evaluations of poorly-performing configurations early, has been discussed in many methods~\cite{swersky2014freeze,domhan2015speeding,baker2017practical,klein2016learning,pmlr-v80-falkner18a,combine_bohb1,bertrand2017hyperparameter,dai2019bayesian}, including Hyperband (HB)~\cite{li2018hyperband} and BOHB~\cite{pmlr-v80-falkner18a}. 
In Section 3, we will describe HB in more detail.
Among them, LCNET-HB~\cite{klein2016learning} utilizes the LCNET that predicts the learning curve of configurations to sample configurations in HB.
In this paper, we explore to use the multi-fidelity quality measurements in the BO framework to further accelerate the HB-style methods.

\noindent
\textbf{Multi-fidelity Optimization} methods exploit the low-fidelity measurements about the objective function $f$ to guide the search for the optimum of $f$, by conducting cheap low-fidelity evaluations proactively, instead of early stopping~\cite{swersky2013multi,kleinfbhh17,kandasamy2017multi,poloczek2017multi,hu2019multi,sen2018noisy,wu2019practical,Wu2019,takeno2020multifidelity}. For instance, FABOLAS~\cite{kleinfbhh17} and TSE~\cite{hu2019multi} evaluate configurations on subsets of the training data and use the generated low-fidelity measurements to infer the quality on the full training set.

\noindent
\textbf{Transfer Learning} methods for HPO aim to take advantage of auxiliary knowledge/information acquired from the past HPO tasks (source problems) to achieve a faster convergence for the current HPO task (target problem)~\cite{bardenet2013collaborative,yogatama2014efficient,schilling2015hyperparameter,wistuba2016two,schilling2016scalable,golovin2017google,feurer2018scalable}. 
While sharing a similar idea, here we investigate to speed up HB-style methods by using the multi-fidelity measurements from \emph{the current HPO task}, instead of the measurements from past similar tasks. Thus, our work is inspired by, but orthogonal to the transfer learning-related methods.

\begin{algorithm}[tb]
  \caption{Pseudo code for Hyperband.}
  \label{algo:outer_lp_hb}
  \textbf{Input}: maximum amount of resource that can be allocated to a single hyperparameter configuration $R$, the discard proportion $\eta$, and hyperparameter space $\mathcal{X}$.
  
  \begin{algorithmic}[1]
  \STATE Initialize $s_{max}=\lfloor log_{\eta}(R) \rfloor$, $B = (s_{max}+1)R$.
  \FOR{$s \in \{s_{max}, s_{max}-1, ..., 0\}$}
      \STATE $n_1 = \lceil \frac{B}{R}\frac{\eta^s}{s+1} \rceil$, $r_1 = R\eta^{-s}$.
      \STATE sample $n_1$ configurations from $\mathcal{X}$ randomly.
      \STATE execute the SH with the $n_1$ configurations and $r_1$ as input (the inner loop).
  \ENDFOR
  \STATE \textbf{return} the configuration with the best evaluation result.
\end{algorithmic}
\end{algorithm}

\begin{figure}[htb]
	\centering
		\scalebox{1.0}[1.0] {
		\includegraphics[width=1\linewidth]{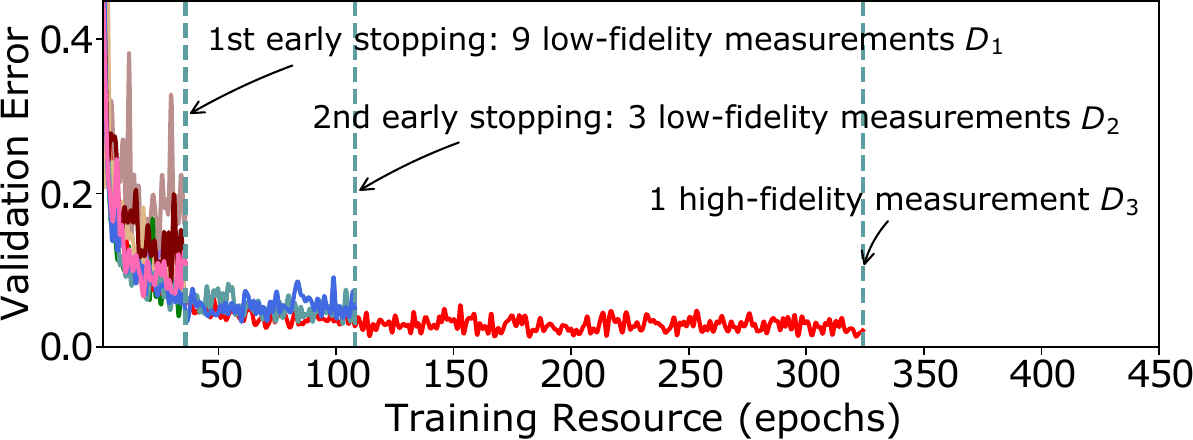}
         }
	\caption{Successive halving process (the inner loop of HB) in tuning a neural network where $n_1=9$, $r_1=1$, $R=9$, and $\eta=3$; one unit of resource corresponds to 36 epochs. First, 9 configurations are evaluated with 1 unit of resource. Then the top 3rd configurations continue their evaluations with 3 units of resources. Finally, only one configuration is evaluated with the maximum training resource $R$.}
    \label{hb_sh}
\end{figure}

\begin{table}[htb]
\LARGE
  \centering
  \resizebox{1\columnwidth}{!}{
  \begin{tabular}{|c|lll|lll|lll|lll|lll|} 
    \hline
    \multirow{1}{*}{} &
    \multicolumn{3}{l|}{$s=4$} &
    \multicolumn{3}{l|}{$s=3$} &
    \multicolumn{3}{l|}{$s=2$} &
    \multicolumn{3}{l|}{$s=1$} &
    \multicolumn{3}{l|}{$s=0$} \cr
    {$i$} & {$n_i$} & & {$r_i$} &
            {$n_i$} & & {$r_i$} &
            {$n_i$} & & {$r_i$} &
            {$n_i$} & & {$r_i$} &
            {$n_i$} & & {$r_i$} \\

    \hline
    {$1$} & {\underline{$81$}} & & {$1$} & 
            {\underline{$27$}} & & {$3$} & 
            {\underline{$9$}} & & {$9$} & 
            {\underline{$6$}} & & {$27$} & 
            {$\bm{5}$} & & $81$ \\
    {$2$} & {\underline{$27$}} & & {$3$} & 
            {\underline{$9$}} & & {$9$} & 
            {\underline{$3$}} & & {$27$} & 
            {$\bm{2}$} & & {$81$} &  & &   \\
    {$3$} & {\underline{$9$}} & & {$9$} & 
            {\underline{$3$}} & & {$27$} &  
            {$\bm{1}$} & & {$81$} & & & & & & \\
    {$4$} & {\underline{$3$}} & & {$27$} & 
            {$\bm{1}$} & & {$81$} & & & & & & & & & \\
    {$5$} & {$\bm{1}$} & & {$81$} & & & & & & & & & & & & \\
    \hline

  \end{tabular}
  }
  \caption{The values of $n_i$ and $r_i$ in the HB evaluations. Here $R = 81$ and $\eta = 3$. Each column displays an inner loop (SH process). The pair ($n_i$, $r_i$) in each cell means there are $n_i$ configuration evaluations with $r_i$ units of training resources.}
  \label{sh_outer_loop}
\end{table}

\section{Bayesian Hyperparameter Optimization and Hyperband}
\label{sec3}
We model the loss $f(\bm{x})$ (e.g., validation error), which reflects the quality of an ML algorithm with the given hyperparameter configuration $\bm{x} \in \mathcal{X}$, as a black-box optimization problem.
The goal of hyperparameter optimization (HPO) is to find $\arg\min_{\bm{x} \in \mathcal{X}}f(\bm{x})$, where the only mode of interaction with the objective function $f$ is to evaluate the given configuration $\bm{x}$. 
Due to the randomness of most ML algorithms, we assume that $f(\bm{x})$ cannot be observed directly but rather through noisy observation $y=f(\bm{x})+\epsilon$, with $\epsilon\sim\mathcal{N}(0, \sigma^2)$.
We now introduce two methods for solving this black-box optimization problem in more detail: Bayesian optimization and Hyperband, which are the basic ingredients in \sys.

\subsubsection{Bayesian Optimization}
The main idea of Bayesian optimization (BO) is as follows. 
Since evaluating the objective function $f$ for a
given configuration $\bm{x}$ is very expensive, it approximates $f$ using a probabilistic surrogate model $M:p(f|D)$ that is much cheaper to evaluate.
Given a configuration $\bm{x}$, the surrogate model $M$ outputs the posterior predictive distribution at $\bm{x}$, that is,
$f(\bm{x}) \sim \mathcal{N}(\mu_{M}(\bm{x}), \sigma^2_{M}(\bm{x}))$.
In the $n^{th}$ iteration, BO methods iterate the following three steps: (1) use the surrogate model $M$ to select a configuration that maximizes the acquisition function $\bm{x}_{n}=\arg\max_{\bm{x} \in \mathcal{X}}a(\bm{x}; M)$, where the acquisition function is used to balance the exploration and exploitation; (2) evaluate the configuration $\bm{x}_{n}$ to get its performance $y_{n}=f(\bm{x}_{n})+\epsilon$ with $\epsilon \sim \mathcal{N}(0, \sigma^2)$; (3) add this measurement $(\bm{x}_{n}, y_{n})$ to the observed quality measurements $D = \{(\bm{x}_1, y_1),...,(\bm{x}_{n-1}, y_{n-1})\}$, and refit the surrogate model on the augmented $D$.
Expected improvement (EI)~\cite{jones1998efficient} is a common acquisition function:
\begin{equation}
\begin{small}
\label{eq_ei}
a(\bm{x}; M)=\int_{-\infty}^{\infty} \max(y^{\ast}-y, 0)p_{M}(y|\bm{x})dy,
\end{small}
\end{equation}
where $y^{\ast}$ is the best observed performance in $D$, i.e., $y^{\ast}=\min\{y_1, ..., y_n\}$, and $M$ is the probabilistic surrogate model. 
By maximizing this EI function $a(\bm{x}; M)$ over the hyperparameter space $\mathcal{X}$, BO methods can find a configuration with the largest EI value to evaluate in each iteration.

\textbf{\emph{Low-efficiency issue}} One fundamental challenge of BO is that, for models that are computationally expensive to train
, each complete evaluation of configuration $\bm{x}$ often takes a significant amount of time. 
Given a limited budget, few measurements
can be obtained, and it is insufficient for BO methods to fit a surrogate that approximates $f$ well.
In this case, BO methods fail to converge to the optimal solution quickly~\cite{wang2013bayesian,li2020efficient}.

\subsubsection{Hyperband}
To accommodate the above issue of BO, Hyperband (HB)~\cite{li2018hyperband} proposes to speed up configuration evaluations by early stopping the badly-performing configurations. 
It has the following two loops: 

\noindent
\emph{\textbf{(1) Inner Loop: successive halving (SH)}} \quad
Given a kind of training resource (e.g., the number of iterations, the size of the  training subset), 
HB first evaluates $n_1$ hyperparameter configurations with the initial $r_1$ units of resources, and ranks them by the evaluation performance.
Then HB continues the top $\eta^{-1}$ configurations with $\eta$ times larger resources (usually $\eta=3$), that's, $n_2=n_1*\eta^{-1}$ and $r_2=r_1*\eta$, and stops the evaluations of the other configurations in advance. 
This process repeats until the maximum training resource $R$ is reached, that's, $r_{i}=R$.
We provide an example to illustrate this procedure in Figure \ref{hb_sh}.

\noindent
\emph{\textbf{(2) Outer Loop: grid search of $n_1$ and $r_1$}} \quad 
Given a fixed budget $B$, the values of $n_1$ and $r_1$ should be carefully chosen because a larger $n_1$ with a small initial training resource $r_1$ may lead to the elimination of good configurations in SH process by mistake.
There is no prior whether we should use a larger $n_1$ with a small initial training resource $r_1$, or a smaller $n_1$ with a larger $r_1$. 
HB addresses this ``$n$ versus $B/n$'' problem by performing a grid search over feasible values of $n_1$ and $r_1$ in the outer loop. 
Algorithm \ref{algo:outer_lp_hb} shows the enumeration of $n_1$ and $r_1$ in Line 3.
Table \ref{sh_outer_loop} lists the number of evaluations and their corresponding training resources in one iteration of HB.
Note that, {\em the HB algorithm can be called multiple times until the HPO budget exhausts}.

\begin{figure*}[htb]
	\centering
	\subfigure[$D_1$: $r_1 = 1$]{
		\scalebox{0.23}[0.23]{
			\includegraphics[width=1\linewidth]{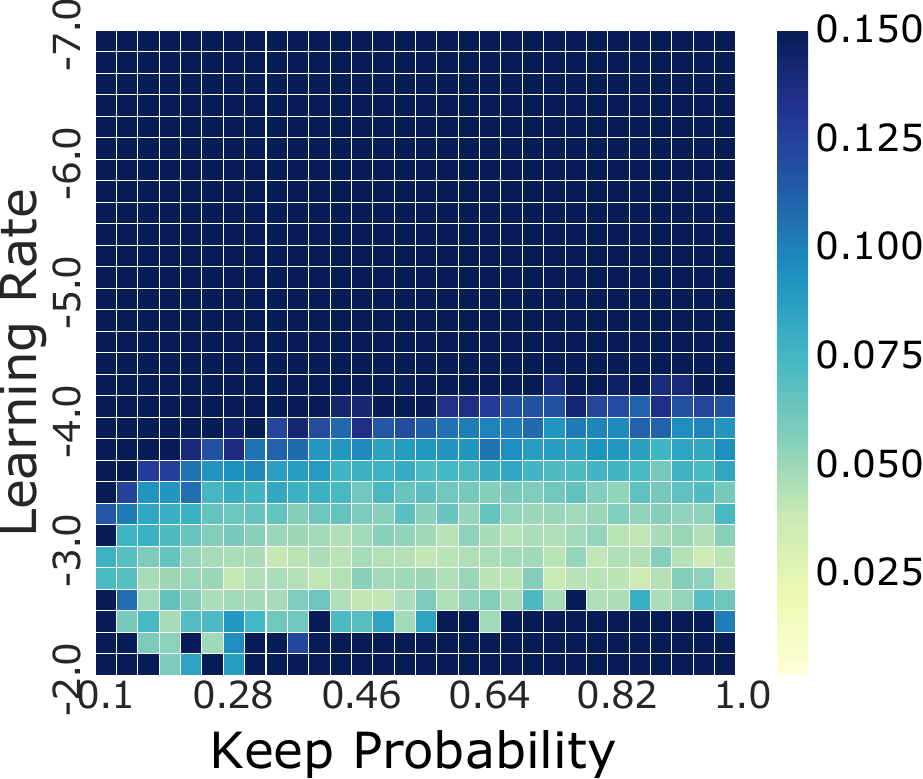}
	}}
	\subfigure[$D_2$: $r_2 = 3$]{
		\scalebox{0.23}[0.23]{
			\includegraphics[width=1\linewidth]{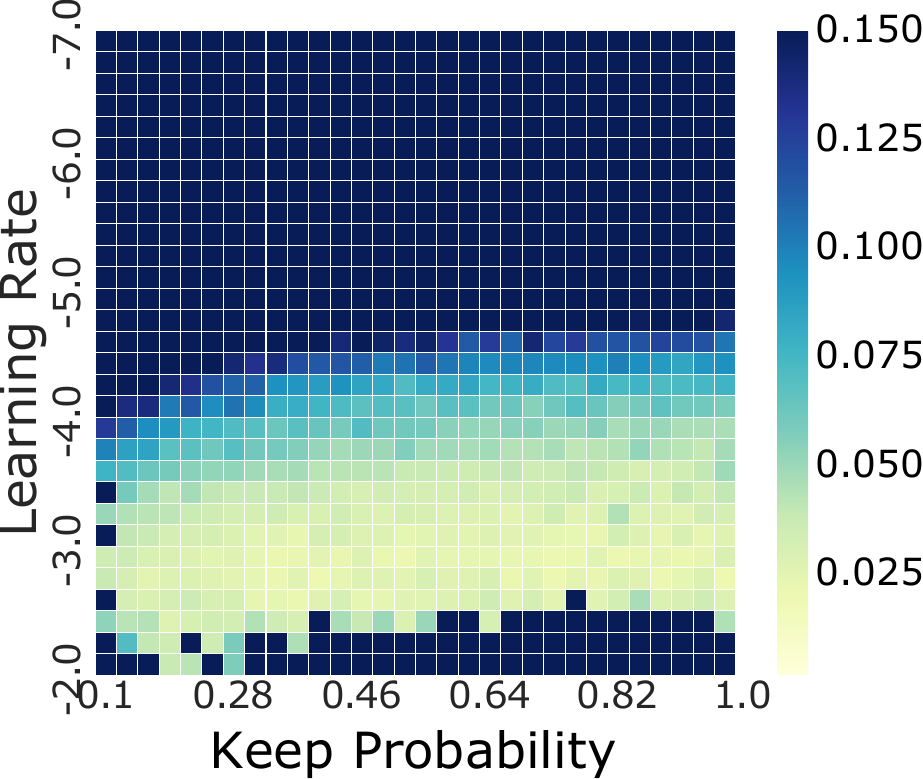}
	}}
    \subfigure[$D_3$: $r_3 = 9$]{
		\scalebox{0.23}[0.23]{
			\includegraphics[width=1\linewidth]{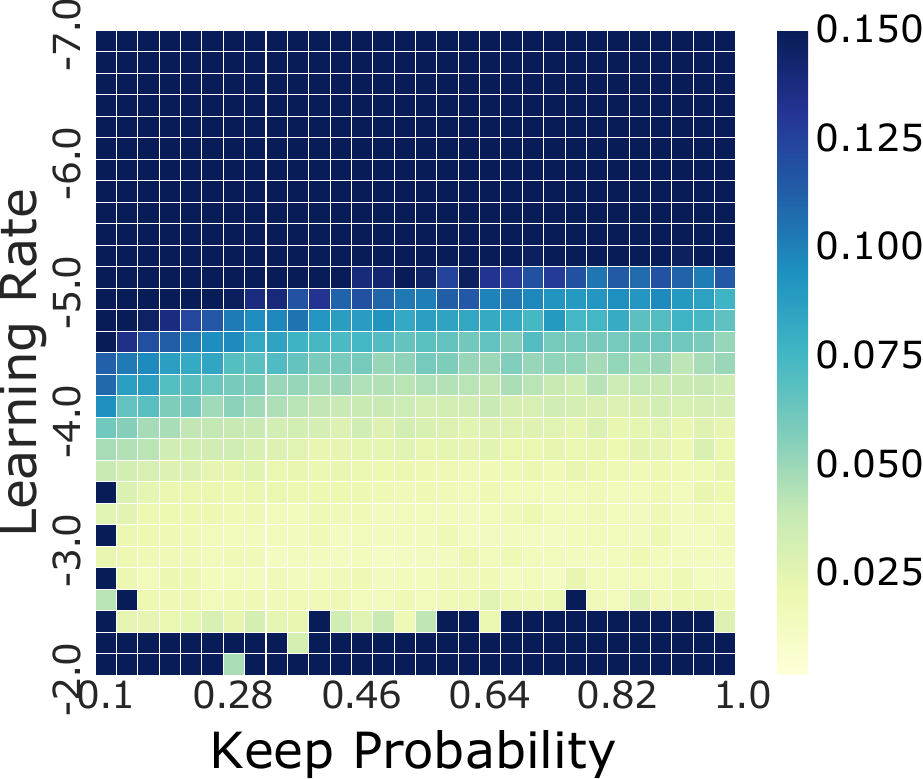}
	}}
    \subfigure[$D_4$: $r_4 = 27$]{
		\scalebox{0.23}[0.23]{
			\includegraphics[width=1\linewidth]{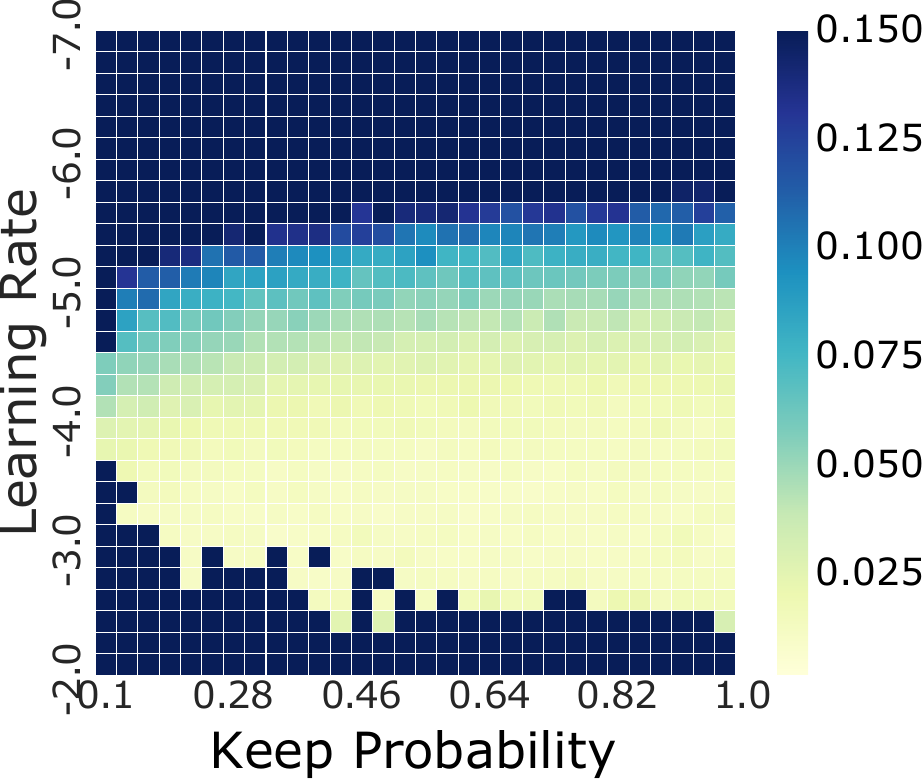}
	}}
	\caption{Validation error of 900 LeNet configurations (30 settings of keep probability $\lambda = 1 - dropout\_value$ in dropout layer and 30 settings of the learning rate $\alpha$ on a base-10 log scale in [-7, -2]) on the MNIST dataset using different training resources $r={1,3,9,27}$. $R = 27$, $K=4$, and one unit of resource corresponds to an epoch.}
  \label{intermediate_distribution}
\end{figure*}

\textbf{The problem of HB and BOHB.}
The disadvantage of HB lies in that it samples configurations randomly from the uniform distribution. 
To improve it, BOHB utilizes a BO component to sample configurations in HB iteration, instead of the uniform distribution. 
However, BOHB also suffers from the low-efficiency issue due to the scarcity of high-fidelity quality measurements. 
Consequently, BOHB does not fully unleash the potential for combining BO and HB.

\section{Efficient Hyperband using Multi-Fidelity Quality Measurements}
\label{sec4}

In this section, we introduce MFES-HB, an efficient Hyperband method that can utilize the multi-fidelity quality measurements in the framework of BO to speed up the convergence of configuration search.
First, we investigate the characteristics of multi-fidelity measurements, and then describe the proposed Multi-Fidelity Ensemble Surrogate (MFES) that is capable of extracting instrumental information from multi-fidelity measurements effectively. 

\subsection{Multi-fidelity Quality Measurements}
According to the number of training resources used by the evaluations in HB, we can categorize the multi-fidelity measurements into $K$ groups: $D_1, ..., D_K$, where $K=\lfloor log_{\eta}(R) \rfloor+1$, and typically $K$ is less than $7$.
The (quality) measurement ($\bm{x}$, $y$) in each group $D_{i}$ with $i\in[1:K]$ is obtained by evaluating $\bm{x}$ with $r_i=\eta^{i-1}$ units of resources.
Thus $D_K$ denotes the high-fidelity measurements from the evaluations with the maximum training resource $r_K=R$, and $D_{1:K-1}$ denote the low-fidelity measurements obtained from the early-stopped evaluations.
Then we discuss the characteristics of $D_{1:K}$ from two aspects:

\noindent
\emph{\textbf{(1) The number of measurements}} Due to successive halving strategy, the number of measurements in $D_i$, i.e., $N_i=|D_i|$, satisfies that $N_1 > N_2 > ... > N_K$. Here, $N_i$ denotes the total number of measurements in each $D_i$ with training resource $r_i$.
Table~\ref{algo:outer_lp_hb} shows the $N_i$s in one iteration of HB, that is, $N_1=81$, $N_2=54$, $N_3=27$, $N_3=15$, and $N_5=10$.

\noindent
\emph{\textbf{(2) The fidelities of measurements}}
The high-fidelity measurements, $D_K$, consist of the unbiased measurements of the objective function $f$.
The other $D_{i}$s, the low-fidelity measurements, are composed of the biased measurements about $f$.
The BO surrogate model $M_{i}$, fitted on $D_i$ with $i<K$, is to model the objective function $f^{i}$ with training resource $r_i$, instead of the true objective function $f=f^{K}$ with the maximum training resource $R$.
Although $f^{1:K-1}$ are different from $f$, they have some correlation.
\emph{
As $i$ increases, the surrogate $M_i$, learned on the measurements $D_i$ with a larger training resource $r_i=\eta^{i-1}$, can offer a higher-fidelity approximation to $f$ because $r$ is closer to $R$. 
}
Figure~\ref{intermediate_distribution} provides a brief example to illustrate the diversity of the measurement fidelity in $D_{1:K}$. 
The quality measurements are visualized as heat maps, where good configurations with low validation errors are marked by the yellow region. 
By comparing the yellow regions in each sub-figure, we can find that, as $i\in[1:3]$ increases, the measurements in $D_i$ with partial training resource $r=\eta^{i-1}$ gradually approach the (unbiased) high-fidelity measurements in $D_K$, where $K=4$.

Hence we can conclude that (1) although $D_{1:K-1}$ includes the biased measurements about $f$, it could still reveal some instrumental information to model $f$; (2) \emph{the group of quality measurements $D_i$ that offers a higher-fidelity approximation to $f$ has a smaller number of measurements $N_i$.}


\subsection {The Proposed Algorithm}
In MFES-HB, we train $K$ base surrogates on $D_{1:K}$ respectively.
(1) $D_K$ offers the highest fidelity when modeling $f$, however, the measurements in $D_K$ are insufficient to train a BO surrogate that describes $f$ well; 
(2) although $D_{1:K-1}$ have a much larger number of quality measurements, the low-fidelity measurements in $D_{1:K-1}$ with biases cannot approximate $f$ accurately.
Thus none of the base surrogates could approximate $f$ well. 
Instead, we propose to combine the base surrogates to obtain a more accurate approximation to $f$.
However, combining the base surrogates is not trivial as we need to integrate the heterogeneous information behind the base surrogates in a reliable and effective way.

Since the performance $y$ in $D_i$s has different numerical ranges, we standardize them by removing the mean and scaling to unit variance respectively.
In BO, the uncertainty prediction of the surrogate $M_i$ at $\bm{x}$ is a Gaussian, i.e., $f^i(\bm{x}) \sim \mathcal{N}(\mu_{M_i}(\bm{x}), \sigma^2_{M_i}(\bm{x}))$. 
For brevity, we use $\mu_i(\bm{x})$ and $\sigma^2_i(\bm{x})$ to denote the mean and variance of predictions from $M_i$.

\begin{algorithm}[tb]
  \small
  \caption{Pseudo code of MFES-HB}
  \label{algo:framework}
  \textbf{Input}: the hyperparameter space $\mathcal{X}$, the total budget to conduct HPO $B_{hpo}$, maximum amount of resource for a hyperparameter configuration $R$, and the discard proportion $\eta$.\\
  \textbf{Output}: the best configuration found.
  
  \begin{algorithmic}[1]
  \STATE initialize: $D_i=\varnothing$ with $i\in[1:K]$, $M_{ens}=None$, $s_{max}=\lfloor log_{\eta}(R) \rfloor$, $B = (s_{max}+1)R$.
  \WHILE{budget $B_{hpo}$ does not exhaust.}
      \FOR{$s \in \{s_{max}, s_{max}-1, ..., 0\}$}
          \STATE $n_1 = \lceil \frac{B}{R}\frac{\eta^s}{s+1} \rceil$, $r_1 = R\eta^{-s}$.
          \STATE sample $n_1$ configurations: $X=Sample(\mathcal{X}, M_{ens}, n_1)$.
          \STATE execute the SH (inner loop) of HB with $X$ and $r_1$ as input, and collect the new multi-fidelity quality measurements:  $D^{'}_{1:K} = SH(X, r_1)$.
          \STATE $D_i= D_i \cup D^{'}_i$ with $i=[1:K]$.
          \STATE update the MFES: $M_{ens} = Build(D_{1:K})$.
      \ENDFOR
  \ENDWHILE
  \STATE \textbf{return} the best configuration $\bm{x}^{\ast}$ in $D_K$.
\end{algorithmic}
\end{algorithm}

\subsubsection{Ensemble Surrogate with gPoE}
To fully utilize the biased yet informative low-fidelity measurements,
we propose the Multi-Fidelity Ensemble Surrogate (MFES) $M_{ens}$ that can integrate the information from all base surrogates to approximate $f$ effectively. 
Concretely, a weight $w_i$ is assigned to each base surrogate $M_i$, which determines the contribution of $M_i$ to the ensemble surrogate $M_{ens}$, where $w_i \in [0, 1]$ and $\sum_{i=1}^{K}w_i = 1$. 
The base surrogate $M_i$, which offers a more accurate approximation to $f$, should own a larger proportion (larger $w_i$) in $M_{ens}$ and vice versa. 
Thus the weights could reflect the influence of the measurements with different fidelities on $M_{ens}$.
Below, we describe the way to combine the base surrogates with weights.

To enable this ensemble surrogate in the BO framework, given a configuration $\bm{x}$, its posterior prediction at $\bm{x}$ should also be a Gaussian, i.e., $f^{ens}(\bm{x}) \sim \mathcal{N}(\mu_{ens}(\bm{x}), \sigma^2_{ens}(\bm{x})).$
To derive the mean and variance, we need to combine the predictions from base surrogates. 
The most straightforward solution is to use $\mu_{ens}(\bm{x})=\sum_iw_i\mu_{i}(\bm{x})$ and $\sigma^2_{ens}(\bm{x})=\sum_iw_i^2\sigma_i^2(\bm{x})$ by assuming that the predictions from base surrogates are independent. 
However, this assumption is contradictory to the fact that these predictive distributions are correlated as discussed before.
Instead, we propose to use the generalized product of experts (gPoE)~\cite{cao2014generalized,hinton1999products} framework to combine the predictions from $M_{1:K}$s. 
The predictive mean and variance of the ensemble surrogate $M_{ens}$ at $\bm{x}$ are given by:
\begin{equation}
\begin{small}
\label{gpoe}
\begin{aligned}
    & \mu_{ens}(\bm{x})=(\sum_i\mu_i(\bm{x})w_i\sigma_i^{-2}(\bm{x}))\sigma_{ens}^2(\bm{x}), \\
    & \sigma_{ens}^2(\bm{x}) = (\sum_iw_i\sigma_i^{-2}(\bm{x}))^{-1},
\end{aligned}
\end{small}
\end{equation}
where $w_i$ is the weight for the $i^{th}$ base surrogate, and it is used to control the influence of individual surrogate $M_i$. 
Using gPoE has the following two advantages: (1) the unified prediction is still a Gaussian;
(2) unreliable predictions are automatically filtered out from the ensemble surrogate.


\begin{algorithm}[tb]
  \small
  \caption{Pseudo code for \emph{Sample} in MFES-HB}
  \label{algo:sample_func}
  \textbf{Input}: the hyperparameter space $\mathcal{X}$, fraction of random configuration $\rho$,  the MFES: $M_{ens}$, the number of random configurations $N_s$ to optimize EI, and evaluation measurements $D_{1:K}$.\\
  \textbf{Output}: next configuration to evaluate.
  \begin{algorithmic}[1]
  \STATE \textbf{if} \emph{rand() $\le \rho$ or $M_{ens}=None$}, then return a random configuration.
  \STATE draw $N_{s}$ configurations randomly, compute their acquisition (EI) values according to the EI criterion in Eq.\ref{eq_ei}, where $M_{ens}$ is used as the surrogate model $M$.
  \STATE \textbf{return} the configuration with the largest EI value.
\end{algorithmic}
\end{algorithm}

\subsubsection{Weight Calculation Method}
In this section, we propose a heuristic method to compute the weight for each base surrogate.
As mentioned above, the value of $w_i$ should be proportional to the performance of $M_i$ when approximating $f$. 
We measure the approximation performance of $M_i$ to $f$ on the high-fidelity measurements $D_K$, by using a pairwise ranking loss.
In HPO, the ranking loss is more reasonable than the mean square error. 
The real value of the prediction does not matter and we only care about the partial orderings over configurations.
We define the ranking loss as the number of misranked pairs:
\begin{equation}
\begin{small}
    \mathcal{L}(M_i) = \sum_{j=1}^{N_K}\sum_{k=1}^{N_K}\mathds{1}((\mu_i(\bm{x}_j) < \mu_i(\bm{x}_k) \oplus (y_j < y_k)),
    \label{rank_loss}
\end{small}
\end{equation}
where $\oplus$ is the exclusive-or operator, $N_K$ is the number of measurements in $D_K$, and $(\bm{x_i}, y_i)$ is the measurement in $D_K$.
Further, for each $M_i$, we can calculate the percentage of the order-preserving pairs by $p_{i}=1 - \frac{\mathcal{L}(M_i;D_K)}{N_{pairs}}$, where $N_{pairs}$ is the number of measurement combination in $D_K$.
Finally, we apply the following weight discrimination operator to obtain the weight $w_i = \frac{p_i^{\theta}}{\sum_{k=1}^Kp_k^{\theta}}$, where $\theta \in \mathbb{N}$ and it is set to $3$ in our experiments.
Due to $p_i \in [0, 1]$, this operator has a discriminative scaling effect on different $p_i$s: (1) further decrease the weight of bad surrogates, and (2) increase the weight of good surrogates. 

For the base surrogates $M_{1:K-1}$, the ranking loss in Eq.\ref{rank_loss} can measure their ability to approximate $f$, i.e., the generalization performance. 
However, for the surrogate $M_K$ trained on $D_K$ directly, this is an estimate of in-sample error and can not reflect generalization. 
To measure $M_K$'s generalization, we adopt the cross-validation strategy.
First, we train $N_K$ leave-one-out surrogates $M_K^{-i}$ on $D_K$ with measurement $(\bm{x}_i,y_i)$ removed. 
Then the ranking loss for $M_K$ can be computed by $\mathcal{L}(M_K) = \sum_{j=1}^{N_K}\sum_{k=1}^{N_K}\mathds{1}((\mu_K^{-j}(\bm{x}_j) < \mu_K^{-j}(\bm{x}_k) \oplus (y_j < y_k))$. In practice, when $n_K$ is larger than $5$, we use 5-fold cross validation to compute $\mathcal{L}(M_K)$.
In the beginning, $w_K$ is set to 0, and $w_i = \frac{1}{K-1}$ with $i \in [1:K-1]$. This means that we utilize more low-fidelity information due to no high-fidelity info available. 
When $|D_K| \ge 3$, the weights are calculated according to the proposed method.

\subsubsection{Putting It All Together}
Algorithm \ref{algo:framework} illustrates the pseudo code of \sys. 
Before executing each SH (the inner loop) of HB, this method utilizes the proposed MFSE to sample $n_1$ configurations (Line 5) according to the \emph{Sample} procedure in Algorithm~\ref{algo:sample_func}. 
After SH ends (Line 6), each $D_i$ is augmented with the new measurements $D^{'}_i$ (Line 7).
Then, the proposed method utilizes $D_{1:K}$ to build the MFES (Line 8).
The function $Build(D_{1:K})$ includes the following three steps: (1) refit each basic surrogate $M_i$ on the augmented $D_i$; (2) calculate the weight $w_i$ for each surrogate; and (3) use gPOE to combine basic surrogates.
Finally, the best configuration in $D_K$ is returned (Line 11).

\subsection{Discussions}

\textbf{Novelty.} MFES-HB is the first method that explores to combine the benefits of both HB and Multi-fidelity Bayesian optimization (MFBO). 
The state-of-the-art BOHB only uses the high-fidelity measurements to guide configuration selection, while it suffers from the low-efficiency issue due to scarce high-fidelity measurements. To alleviate this issue, we propose to utilize massive low-fidelity measurements. 
However, utilizing the massive and cheap low-fidelity measurements in an effective and efficient manner is not trivial, and we need to balance the ``fidelities and \#measurements'' trade-off introduced in Sec.3.
Further, we propose to build an ensemble surrogate, which can leverage the useful information from the multi-fidelity measurements to guide the configuration search. 
\newline
\textbf{Convergence Analysis.} (1) When the high-fidelity measurements become sufficient, i.e., $n_K$ is larger than a threshold, $w_K$ will be set to $1$ in \sys.
Therefore, the convergence result of \sys will be no worse than the state-of-the-art BOHB's. 
(2) MFES-HB also samples a constant fraction $\rho$ of the configurations randomly (Line 1 in Algorithm~\ref{algo:sample_func}), thus the theoretical guarantee of HB still holds in MFSE-HB.
We provide a detailed analysis of the two guarantees in Appendix A.1 of supplemental materials.

POGPE~\cite{schilling2016scalable} also uses a similar product of Gaussian Process (GP) experts to combine the GP-based surrogates. 
It is trained on the measurements from the past HPO tasks, while the measurements in \sys are obtained from the current HPO task. 
Moreover, the weight in POGPE is set to a constant $w_i=\frac{1}{K}$. 
Multi-fidelity Bayesian optimization (MFBO) methods can accelerate HPO by conducting low-fidelity evaluations proactively. 
However, since most MFBO methods~\cite{swersky2013multi,kleinfbhh17} use the GP in the surrogate model, (1) they cannot support complex configuration spaces easily.
(2) Most MF based methods only support a particular type of training resource as they often rely on some customized optimization structures.
MFES-HB, which inherits the advantages from HB, can support all resource types, e.g., the number of iterations (epochs) for an iterative algorithm, the size of dataset subset, the number of steps in an MCMC chain, etc. 
While many MFBO methods are designed to work on one kind of training resource, e.g., FABOLAS and TSE only support (dataset) subsets as resources, and LCNET-HB only supports epochs (\#iterations) as resources.
(3) These methods are intrinsically sequential and difficult to parallelize. MFES-HB does not have the above three limitations by 1) using a probabilistic random forest-based surrogate and 2) 
inheriting the easy-to-parallel merit from HB. 
In the following section, we evaluate the proposed method and discuss more advantages of \sys.

\section{Experiments and Results} 
\label{section:exp}
To evaluate the proposed \sys, we apply it to tune hyperparameters on several real-world AutoML tasks.
Compared with other methods, four main insights about MFES-HB that we should investigate are as follows:
\begin{small}
\begin{itemize}
    \item Using low-fidelity quality measurements brings benefits.
    
    \item The proposed MFES can effectively utilize the multi-fidelity quality measurements from HB evaluations.
    
    \item \sys can greatly accelerate HPO tasks. It reaches a similar performance like other methods but spends much less time.
    
    \item \sys has the following advantages: scalability, generality, flexibility, practicability in AutoML systems.
\end{itemize}
\end{small}

\begin{table}[tb]
\linespread{1.3}
\fontsize{41pt}{41pt}\selectfont
  \centering
  \resizebox{1.0\columnwidth}{!}{
  \begin{tabular}{|l|l|c|l|c|l|l|}
    \hline
    Task & Datasets & \scalebox{1.68}{$|\mathcal{X}|$} & \scalebox{1.68}{$R$} & \scalebox{1.68}{$B_{hpo}$} & Type & Unit \\
    \hline
    FCNet & MNIST & 10 & 81 & 5h & \#Iterations & 0.5 epoch\\
    ResNet & CIFAR-10 & 6 & 81 & 28h & \#Iterations & 2 epochs \\
    XGBoost & Covtype & 8 & 27 & 7.5h & Data Subset & 1/27 \#samples \\
    \hline
    AutoML & 10 Datasets & 110 & 27 & 4h & Data Subset & 1/27 \#samples \\
    \hline
  \end{tabular}
  }
  \caption{Four real-world HPO tasks. $|\mathcal{X}|$ is the number of hyperparameters in $
\mathcal{X}$; $R$ is the maximum training resource; $B_{hpo}$ is the total HPO budget.}
  \label{exp_tasks}
\end{table}

\begin{figure*}[htb]
	\centering
	\subfigure[Feasibility analysis]{
		\scalebox{0.32}{
			\includegraphics[width=1\linewidth]{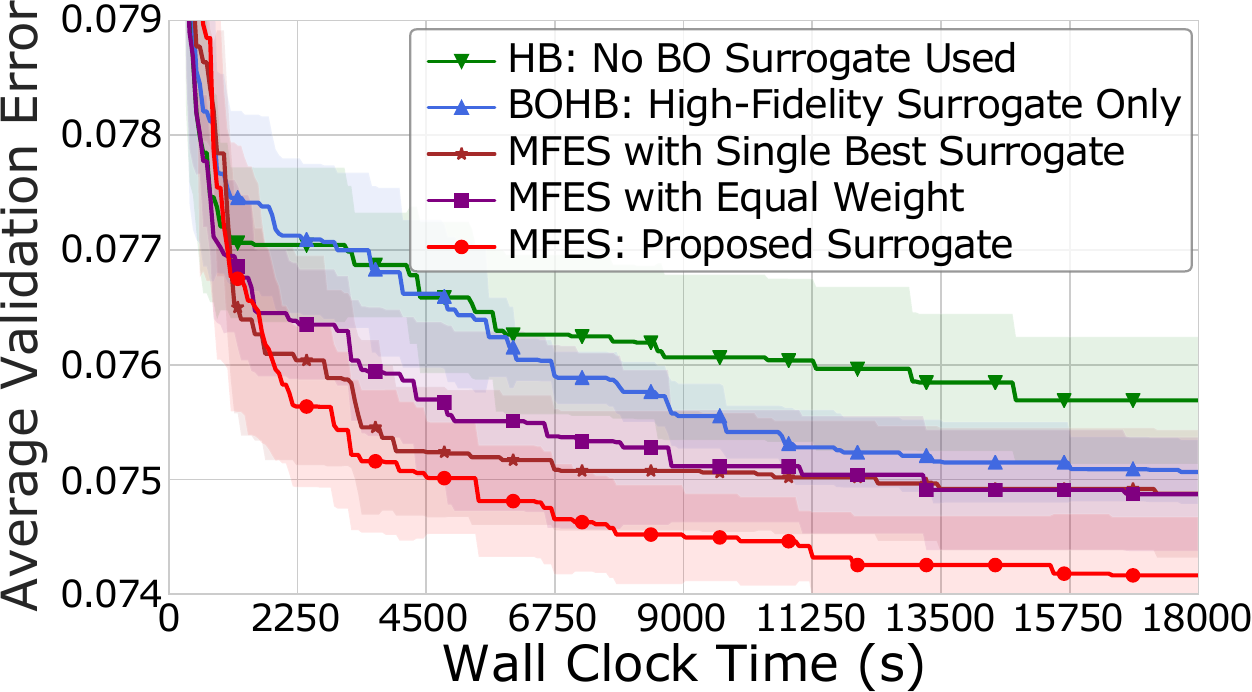}
	}}
	\subfigure[Weight update]{
		\scalebox{0.32}{
			\includegraphics[width=1\linewidth]{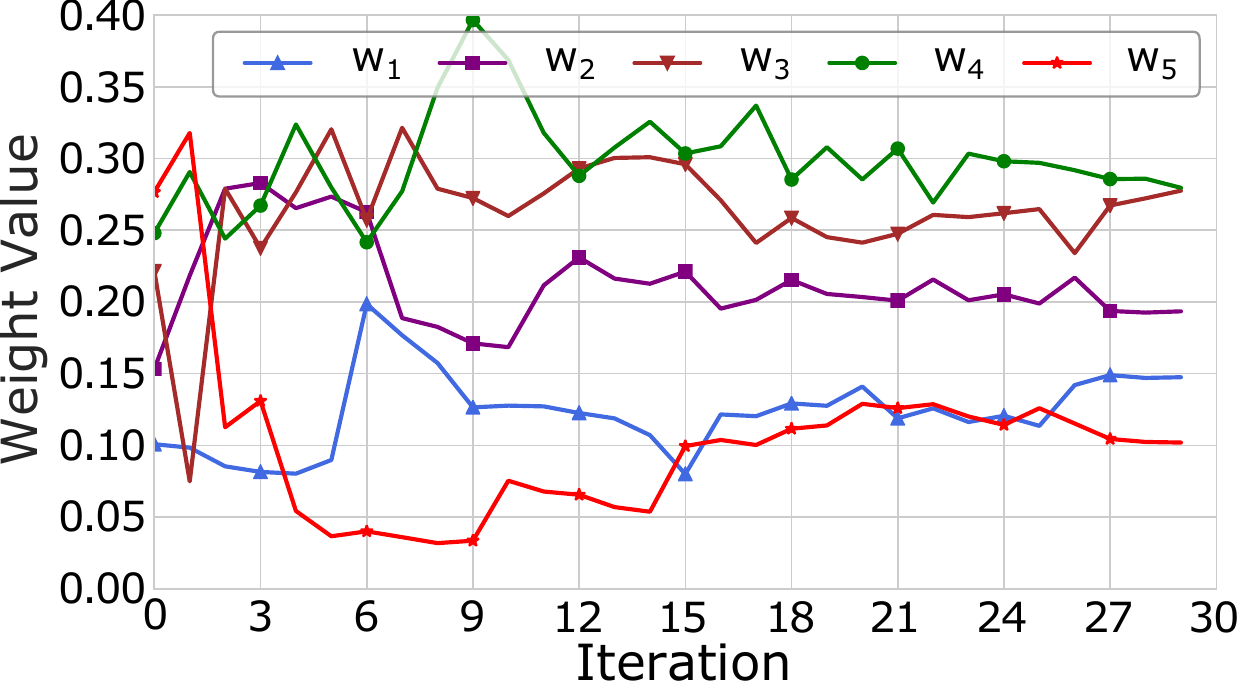}
	}}
    \subfigure[gPoE \& Parameter check]{
		\scalebox{0.32}{
			\includegraphics[width=1\linewidth]{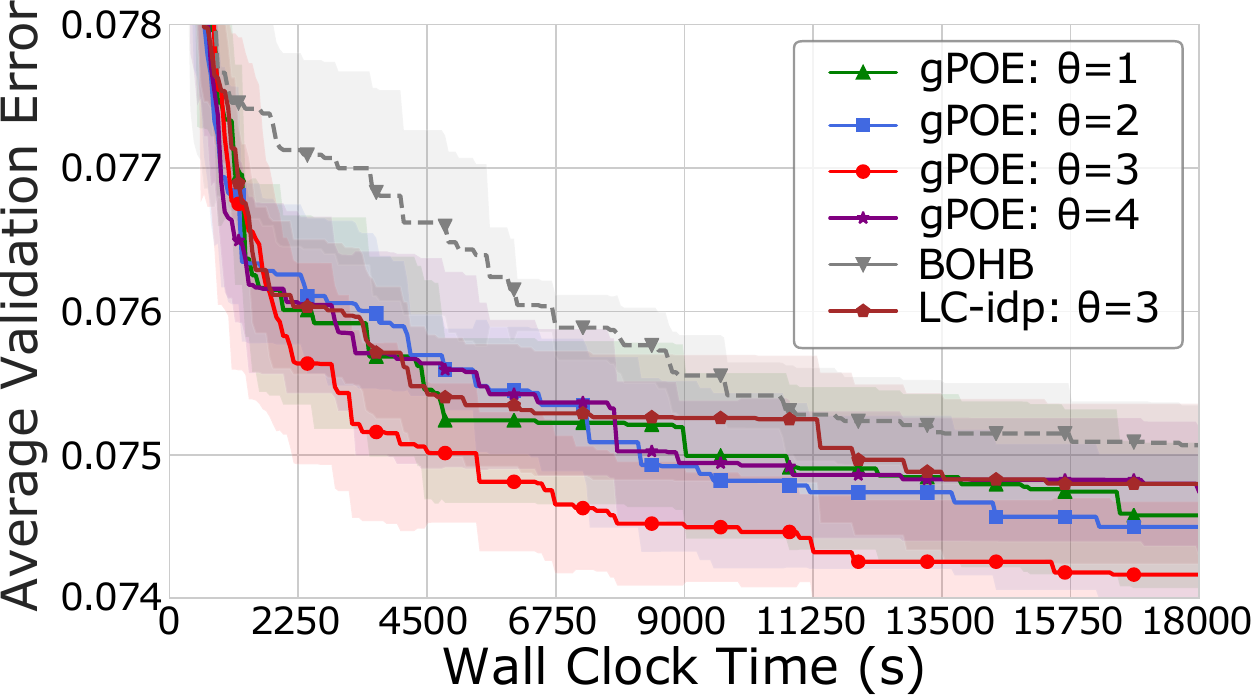}
	}}
	\caption{Optimizing $10$ hyperparameters of FCNet on MNIST.}
  \label{fc_evaluation_exp}
\end{figure*}

\begin{figure*}[htb]
	\centering
	\subfigure[Results on FCNet]{
  	\label{exp:fcnet_automl1}
		\scalebox{0.32}{
			\includegraphics[width=1\linewidth]{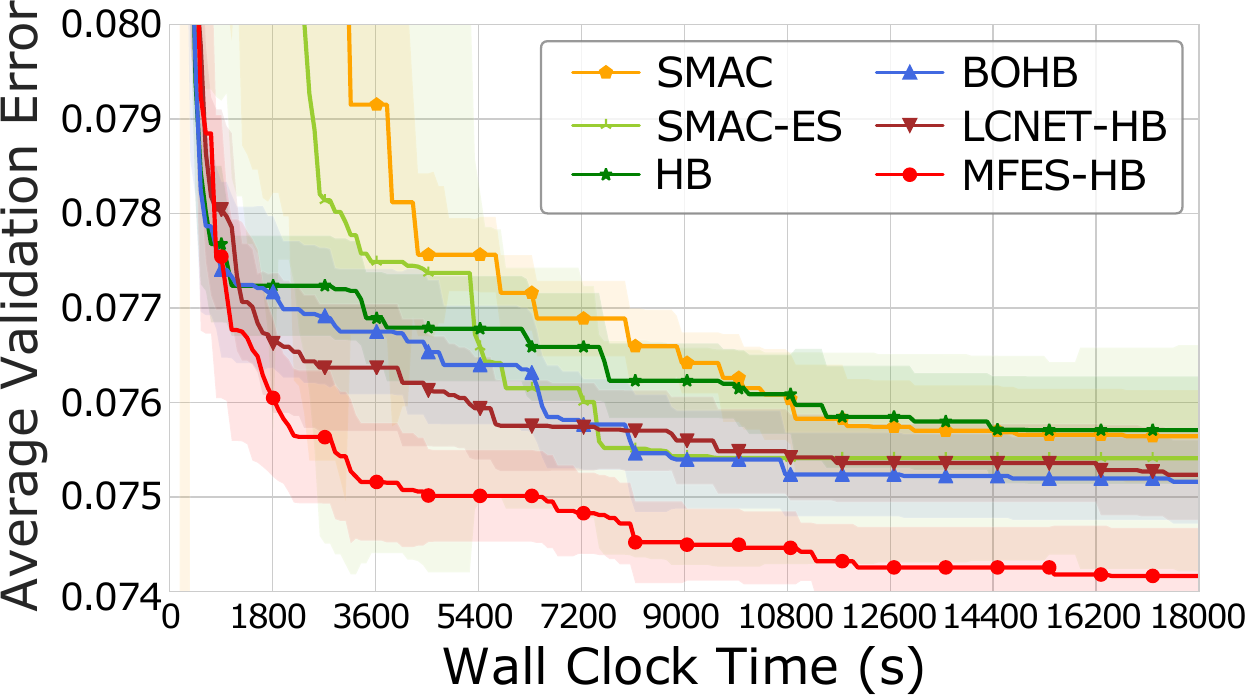}
	}}
    \subfigure[AutoML]{
	\label{exp:fcnet_automl2}
		\scalebox{0.32}{
			\includegraphics[width=1\linewidth]{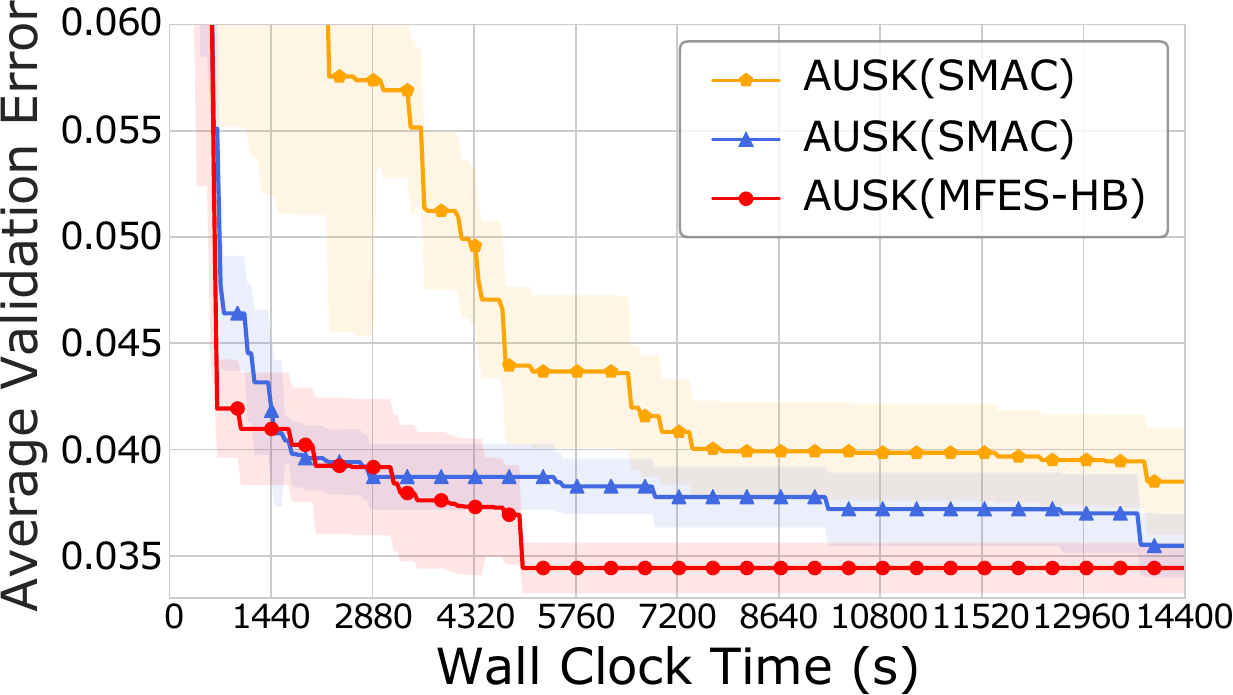}
	}}
	\caption{Results for optimizing FCNet on MNIST (sequential) and AutoML on Letter (parallel).}
  \label{exp:fcnet_automl}
\end{figure*}

\begin{figure*}[htb]
	\centering
	\subfigure[FCNet]{
		\scalebox{0.32}{
			\includegraphics[width=1\linewidth]{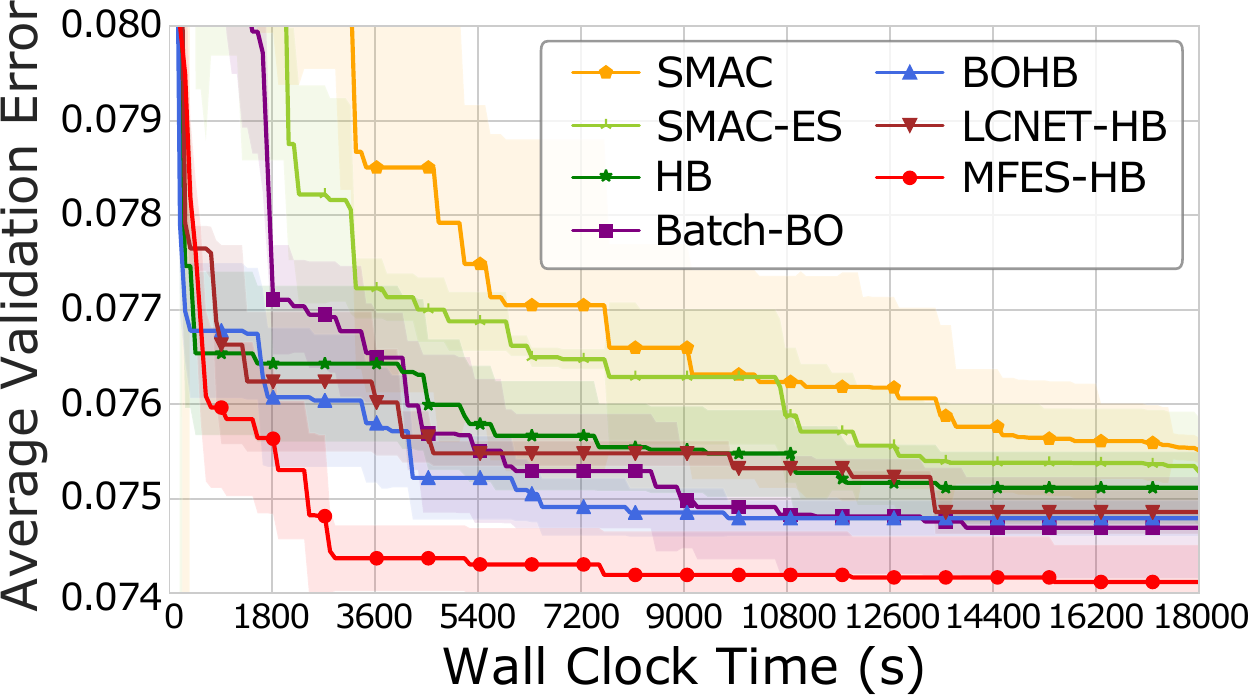}
	}}
	\subfigure[ResNet]{
		\scalebox{0.32}{
			\includegraphics[width=1\linewidth]{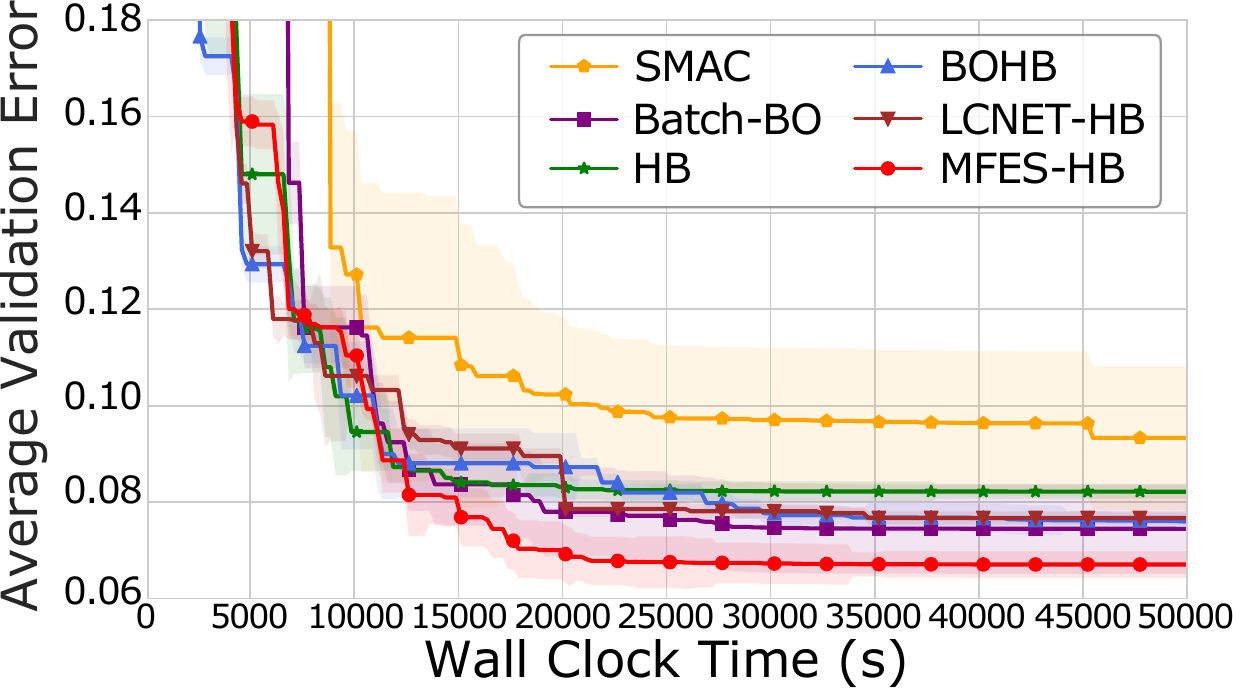}
	}}
	\subfigure[XGBoost]{
		\scalebox{0.32}{
			\includegraphics[width=1\linewidth]{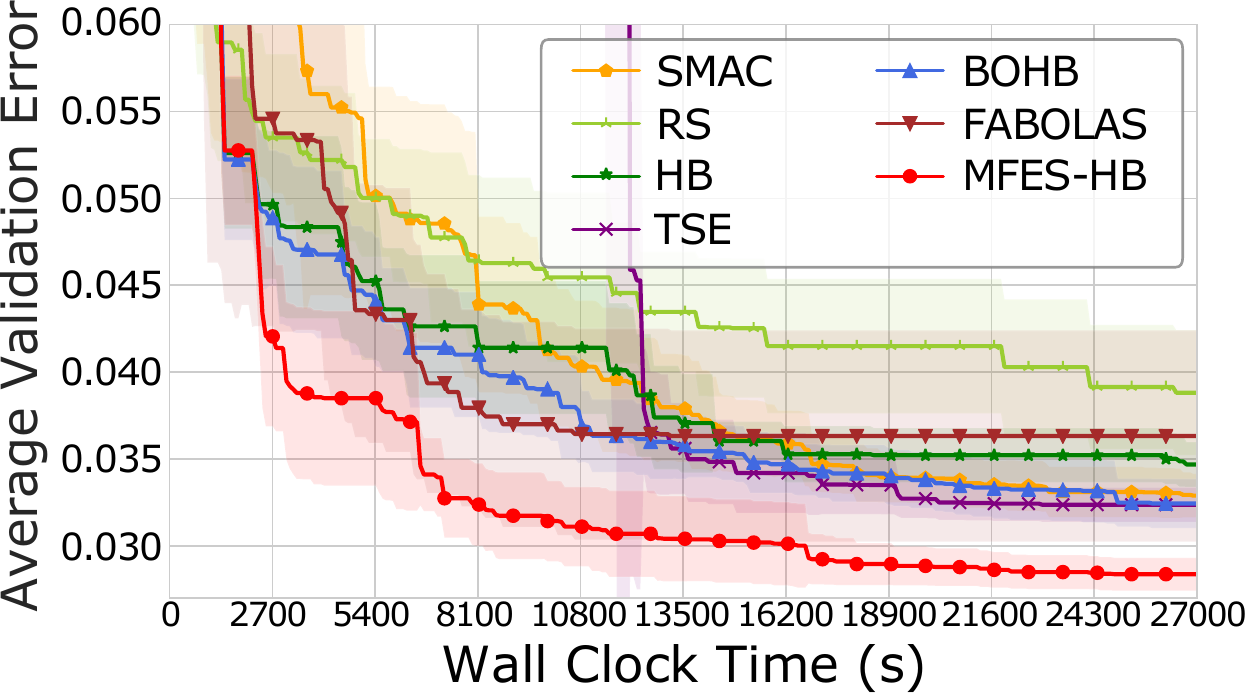}
	}}
	\caption{Results for optimizing FCNet on MNIST, ResNet on CIFAR-10, and XGBoost on Covtype.}
  \label{parallel_setting}
\end{figure*}

\subsection{Experiment Settings}
\textbf{Baselines.} We compared \sys with the following eight baselines: (1) HB: the original Hyperband~\cite{li2018hyperband},
(2) BOHB~\cite{pmlr-v80-falkner18a}: a model-based Hyperband that uses TPE-based surrogate fitted on the high-fidelity measurements to sample configurations,  
(3) LCNET-HB~\cite{kleinfbhh17}: also a model-based Hyperband, it utilizes the LCNet that predicts the learning curve of configurations to sample configurations in HB, 
(4) SMAC~\cite{hutter2011sequential}: a widely-used BO method with high-fidelity evaluations, 
(5) SMAC-ES~\cite{domhan2015speeding}: a variant of SMAC with the learning curve extrapolation-based early stopping,
(6) Batch-BO~\cite{gonzalez2016batch}: a parallel BO method that conducts a batch of high-fidelity evaluations concurrently.
(7) FABOLAS~\cite{klein2016learning} and (8) TSE~\cite{hu2019multi}: the multi-fidelity optimization methods that utilize the cheap fidelities of $f$ on subsets of the training data.
Besides the two compared MF methods (FABOLAS and TSE), we also considered the following 3 methods: MF-MES~\cite{takeno2020multifidelity}, taKG~\cite{Wu2019} and MTBO~\cite{swersky2013multi}.

\noindent
\textbf{HPO Tasks.}
Table~\ref{exp_tasks} describes 4 real-world AutoML HPO tasks in our experiments. 
For example, in FCNet, we optimized $10$ hyperparameters on MNIST; the resource type is the number of iterations; one unit of resource corresponds to 0.5 epoch, and the total HPO budget $B_{hpo}$ for each baseline is $5$ hours.
In addition, to investigate the practicability and the performance of \sys in AutoML systems, we also implemented \sys in Auto-Sklearn (AUSK)~\cite{feurer2015efficient}, and compared it with the built-in HPO method --- SMAC in AUSK, on 10 public datasets.
More details about the configuration space $\mathcal{X}$ and datasets (12 in total) can be found in Appendix A.2 and A.3 respectively.

\noindent
\textbf{Dataset Split, Metric and Parameter Setting.}
In each experiment, we randomly divided 20\% of the training dataset as the validation set, tracked the wall clock time (including optimization overhead and evaluation cost), and stored the lowest validation error after each evaluation.
All methods are repeated 10 times with different random seeds, and the mean($\pm$std) validation errors across runs are plotted.
Moreover, the best models found by the baselines are applied to the test dataset, and test errors are reported.
All methods are discussed based on two metrics: (1) the time taken for reaching the same validation error, (2) the test performance.
As recommended by HB and BOHB, $\eta$ is set to $3$ for the HB-based methods, and $\rho=0.2$.
In \sys, we implemented the MFES based on the probabilistic random forest from the SMAC package; the parameter $\theta$ used in weight discrimination operator is set to $3$.
Figure~\ref{fc_evaluation_exp} (c) depicts the sensitivity analysis about $\theta$.
The same parallel mechanism in BOHB is adopted in the parallel experiments.
More details about the experiment settings, hardware, the parameter settings of baselines, and their implementations (including source code) are available in Appendix A.4, A.5, and A.6.

\subsection{Empirical Analysis}
Figures~\ref{fc_evaluation_exp} and ~\ref{exp:fcnet_automl1} illustrate the results on FCNet, where we investigated (1) the feasibility of the low-fidelity measurements and (2) the effectiveness of MFES.
Figures~\ref{parallel_setting} and ~\ref{exp:fcnet_automl2} show the results on four HPO tasks, where we studied the efficiency of \sys.
Table~\ref{table_res} lists the test results.
Below, we will verify four insights based on these results.

\textbf{Using low-fidelity quality measurements brings benefits.} 
Figure~\ref{fc_evaluation_exp} (a) shows the results of (1) the different versions of MFES that utilize the multi-fidelity measurements and (2) BOHB that uses the high-fidelity measurements only, on FCNet.
MFES with single best surrogate means that it uses the base surrogate with the smallest ranking loss defined in Eq.\ref{rank_loss} to sample configurations in HB.
MFES with equal weight means that, for each surrogate $M_i$, $w_i=\frac{1}{K}$.
MFES refers to the proposed ensemble surrogate with a ranking loss based weight calculation method.
We can observe that using low-fidelity measurements can bring benefits to achieve a faster convergence in HPO. 

\textbf{MFES can exploit the multi-fidelity measurements effectively. }
From Figure~\ref{fc_evaluation_exp}(a), we can find that two variants (MFES with the single best surrogate and MFES with equal weight) cannot beat the proposed MFSE method using gPoE.
As shown in Figure~\ref{fc_evaluation_exp}(c), gPOE is more reasonable and effective in combining the base surrogates than the linear combination under the independent assumption (LC-idp curve).
Based on the above results, the proposed MFSE, which combines base surrogates using gPOE with the ranking loss based weight calculation technique, is an effective surrogate to utilize the multi-fidelity measurements.
To further investigate the weight update process, the values of $w_i$s across iterations are illustrated in Figure~\ref{fc_evaluation_exp}(b).
The surrogates trained on the lowest-fidelity measurements and the scarce high-fidelity measurements have relatively smaller weights; the surrogates with medium-fidelity measurements own larger weights. 
Figure~\ref{fc_evaluation_exp}(c) depicts the parameter sensitivity check of $\theta$.
Finally, Figure~\ref{exp:fcnet_automl1} shows the HPO results of all methods on FCNet, and \sys obtains a (more than) $5\times$ speedup over the baselines.

\begin{table}[tb]
\renewcommand{\thesubtable}{\footnotesize(\alph{subtable})}
\linespread{0.95}
\centering
\setlength{\tabcolsep}{5mm}{
\subtable[\footnotesize Test results of baselines]{
\small
\begin{tabular}{lcccccc}
    \toprule
    Method & FCNet & ResNet & XGB  \\ 
    \midrule
    SMAC & 7.63      & 9.10  & 3.52  \\
    SMAC (ES) & 7.49 & 8.37  & -      \\
    Batch BO & 7.47  & 7.98  & 3.00  \\
    HB & 7.55        & 8.40  & 3.56  \\
    LCNET-HB & 7.49  & 8.21  & -     \\
    BOHB & 7.48      & 8.10  & 3.16  \\
    FABOLAS & -      & -     & 3.40  \\
    TSE & -          & -     & 3.12  \\
    \sys & \textbf{7.38} & \textbf{7.49} & \textbf{2.65} \\
    \bottomrule
\end{tabular}
}
}
\\
\setlength{\tabcolsep}{6mm}{
\subtable[\footnotesize Results on 10 AutoML datasets]{
\small
\begin{tabular}{lcc}
     \toprule
     Dataset & AUSK & AUSK(new) \\
     \midrule
     MNIST & 3.39  & \textbf{2.15} \\
     Letter & 3.85 & \textbf{3.44} \\
     Higgs & 26.84 & \textbf{26.79} \\
     Electricity & \underline{6.18} & \underline{6.21} \\
     Kropt & 19.84 & \textbf{13.08} \\
     Mv & 0.03 & \textbf{0.01} \\
     Poker & 12.91 & \textbf{4.30} \\
     Fried & \underline{6.60} & \underline{6.62} \\
     A9a & 17.23 & \textbf{17.09} \\
     2dplanes & 6.59 & \textbf{6.41} \\
     \bottomrule
\end{tabular}
}
}
\caption {Mean test errors (\%) of compared baselines. In Table(a), since SMAC (ES) \& LCNET-HB depend on training iteration and FABOLAS \& TSE only work on sample size, `-' means the invalid cases. In Table(b), AUSK(new) represents Auto-Sklearn using \sys as its HPO optimizer.}
\label{table_res}
\end{table}

\begin{table}[tb]
\renewcommand{\thesubtable}{\huge(\alph{subtable})}
\linespread{1.1}
\huge
\centering
\resizebox{1\columnwidth}{!}{
\subtable[\huge Results on FCNet]{
\begin{tabular}{lcc}
    \toprule
    Method & Error & Speedups \\ 
    \midrule
    HB & 7.53 & 1.0x    \\
    BOHB & 7.48 & 3.05x \\
    MTBO & 7.50 & 1.07x \\
    MF-MES & 7.45 & 5.7x  \\
    taKG & 7.46 & 4.8x    \\
    MFES-HB & \textbf{7.41} & \textbf{10.1x} \\
    \bottomrule
\end{tabular}
}
\subtable[\huge Results on XGBoost]{
\begin{tabular}{lcc}
     \toprule
     Method & Error & Speedups \\
     \midrule
     HB & 3.48 & 1.0x \\
     BOHB & 3.26 & 1.8x \\
     MTBO & 3.46 & F \\
     MF-MES & 3.11 & 3.1x \\
     taKG & 3.16 & 2.9x \\
     MFES-HB & \textbf{2.97} & \textbf{4.5x} \\
     \bottomrule
\end{tabular}
}
}
\caption {Speedup result over Hyperband (HB) on two benchmarks: FCNet and XGBoost. `F' means the method fails to reach the result of HB.}
\label{table_res1}
\end{table}

\textbf{\sys can accelerate HPO tasks.}
Figures~\ref{parallel_setting} and ~\ref{exp:fcnet_automl2} depict the empirical results on four HPO tasks, where the tasks on FCNet and ResNet are conducted in parallel settings.
\sys spends less time than the compared methods to obtain a sub-optimal performance.
Concretely, \sys achieves the validation error of 7.5\% on FCNet within 0.75 hours, 7.3\% on ResNet within 4.3 hours, and 3.5\% on XGBoost within 2.25 hours.
To reach the same results, it takes other methods about 5 hours on FCNet, 13.9 hours on ResNet, and 7.5 hours on XGBoost.
\sys achieves the $3.2$ to $6.7\times$ speedups in finding a similar configuration.
Particularly, \sys achieves $4.05$ to $10.1\times$ speedups over Hyperband, and $3.3$ to $8.9\times$ speedups over the state-of-the-art BOHB.
Moreover, Table~\ref{table_res} (a) shows that the configuration found by \sys gets the best test performance.
In addition, when comparing MF-MES, taKG, and MTBO, the final results and speedups over HB in achieving the same final error of HB are reported in Table~\ref{table_res1}.
MFES-HB outperforms these methods, and we see that MFES-HB, which combines the benefits of HB and MFBO, works well.
Therefore, this demonstrates that \sys can conduct HPO efficiently and effectively.

\textbf{The advantages of \sys.}
\sys can easily handle HPO problems with $6$ to $110$ hyperparameters (scalability).
Particularly, the AutoML task on 10 datasets involves a very high-dimensional space: 110 hyperparameters in total.
In addition, \sys supports (1) complex hyperparameter space by using the BO component from SMAC (generality), and (2) HPO with different resource types (flexibility);
while most multi-fidelity optimization approaches only support one type of resource, and cannot be extended to the other types easily.
Finally, on 10 datasets, we compared the performance of \sys with the built-in HPO algorithm (SMAC) in Auto-Sklearn (AUSK).
Figure~\ref{exp:fcnet_automl2} shows the results on dataset Letter, and Table~\ref{table_res} (b) demonstrates its practicability and effectiveness in AutoML system.

\section{Conclusion}
In this paper, we introduced \sys, an efficient Hyperband method that utilizes multi-fidelity quality measurements to accelerate HPO tasks.
The multi-fidelity ensemble surrogate is proposed to integrate quality measurements with multiple fidelities effectively.
We evaluated \sys on a wide range of AutoML HPO tasks, and demonstrated its superiority over the competitive approaches.

\section{Acknowledgments}
This work is supported by the National Key Research and Development Program of China (No.2018YFB1004403), NSFC (No.61832001, 61702015, 61702016), Beijing Academy of Artificial Intelligence (BAAI), and Kuaishou-PKU joint program. Bin Cui is the corresponding author.

\bibliography{reference}

\begin{thebibliography}{46}
\providecommand{\natexlab}[1]{#1}
\providecommand{\url}[1]{\texttt{#1}}
\providecommand{\urlprefix}{URL }
\expandafter\ifx\csname urlstyle\endcsname\relax
  \providecommand{\doi}[1]{doi:\discretionary{}{}{}#1}\else
  \providecommand{\doi}{doi:\discretionary{}{}{}\begingroup
  \urlstyle{rm}\Url}\fi

\bibitem[{Baker et~al.(2017)Baker, Gupta, Raskar, and
  Naik}]{baker2017practical}
Baker, B.; Gupta, O.; Raskar, R.; and Naik, N. 2017.
\newblock Practical neural network performance prediction for early stopping.
\newblock \emph{arXiv preprint arXiv:1705.10823} 2(3): 6.

\bibitem[{Bardenet et~al.(2013)Bardenet, Brendel, K{\'e}gl, and
  Sebag}]{bardenet2013collaborative}
Bardenet, R.; Brendel, M.; K{\'e}gl, B.; and Sebag, M. 2013.
\newblock Collaborative hyperparameter tuning.
\newblock In \emph{International Conference on Machine Learning}, 199--207.

\bibitem[{Bergstra et~al.(2011)Bergstra, Bardenet, Bengio, and
  K{\'e}gl}]{bergstra2011algorithms}
Bergstra, J.~S.; Bardenet, R.; Bengio, Y.; and K{\'e}gl, B. 2011.
\newblock Algorithms for hyper-parameter optimization.
\newblock In \emph{Advances in neural information processing systems},
  2546--2554.

\bibitem[{Bertrand et~al.(2017)Bertrand, Ardon, Perrot, and
  Bloch}]{bertrand2017hyperparameter}
Bertrand, H.; Ardon, R.; Perrot, M.; and Bloch, I. 2017.
\newblock Hyperparameter optimization of deep neural networks: Combining
  hyperband with Bayesian model selection.
\newblock In \emph{Conf{\'e}rence sur l’Apprentissage Automatique}.

\bibitem[{Cao and Fleet(2014)}]{cao2014generalized}
Cao, Y.; and Fleet, D.~J. 2014.
\newblock Generalized product of experts for automatic and principled fusion of
  Gaussian process predictions.
\newblock \emph{arXiv preprint arXiv:1410.7827} .

\bibitem[{Dai et~al.(2019)Dai, Yu, Low, and Jaillet}]{dai2019bayesian}
Dai, Z.; Yu, H.; Low, B. K.~H.; and Jaillet, P. 2019.
\newblock Bayesian Optimization Meets Bayesian Optimal Stopping 1496--1506.

\bibitem[{Domhan, Springenberg, and Hutter(2015)}]{domhan2015speeding}
Domhan, T.; Springenberg, J.~T.; and Hutter, F. 2015.
\newblock Speeding Up Automatic Hyperparameter Optimization of Deep Neural
  Networks by Extrapolation of Learning Curves.
\newblock In \emph{IJCAI}, volume~15, 3460--8.

\bibitem[{Eggensperger et~al.(2013)Eggensperger, Feurer, Hutter, Bergstra,
  Snoek, Hoos, and Leyton-Brown}]{eggensperger2013towards}
Eggensperger, K.; Feurer, M.; Hutter, F.; Bergstra, J.; Snoek, J.; Hoos, H.;
  and Leyton-Brown, K. 2013.
\newblock Towards an empirical foundation for assessing bayesian optimization
  of hyperparameters.
\newblock In \emph{NIPS workshop on Bayesian Optimization in Theory and
  Practice}, volume~10, 3.

\bibitem[{Falkner, Klein, and Hutter(2018)}]{pmlr-v80-falkner18a}
Falkner, S.; Klein, A.; and Hutter, F. 2018.
\newblock {BOHB}: Robust and Efficient Hyperparameter Optimization at Scale.
\newblock In \emph{Proceedings of the 35th International Conference on Machine
  Learning}, 1436--1445.

\bibitem[{Feurer et~al.(2015)Feurer, Klein, Eggensperger, Springenberg, Blum,
  and Hutter}]{feurer2015efficient}
Feurer, M.; Klein, A.; Eggensperger, K.; Springenberg, J.; Blum, M.; and
  Hutter, F. 2015.
\newblock Efficient and robust automated machine learning.
\newblock In \emph{Advances in neural information processing systems},
  2962--2970.

\bibitem[{Feurer, Letham, and Bakshy(2018)}]{feurer2018scalable}
Feurer, M.; Letham, B.; and Bakshy, E. 2018.
\newblock Scalable meta-learning for bayesian optimization using
  ranking-weighted gaussian process ensembles.
\newblock In \emph{AutoML Workshop at ICML}.

\bibitem[{Golovin et~al.(2017)Golovin, Solnik, Moitra, Kochanski, Karro, and
  Sculley}]{golovin2017google}
Golovin, D.; Solnik, B.; Moitra, S.; Kochanski, G.; Karro, J.; and Sculley, D.
  2017.
\newblock Google vizier: A service for black-box optimization.
\newblock In \emph{Proceedings of the 23rd ACM SIGKDD International Conference
  on Knowledge Discovery and Data Mining}, 1487--1495. ACM.

\bibitem[{Gonz{\'a}lez et~al.(2016)Gonz{\'a}lez, Dai, Hennig, and
  Lawrence}]{gonzalez2016batch}
Gonz{\'a}lez, J.; Dai, Z.; Hennig, P.; and Lawrence, N. 2016.
\newblock Batch bayesian optimization via local penalization.
\newblock In \emph{Artificial Intelligence and Statistics}, 648--657.

\bibitem[{Goodfellow, Bengio, and Courville(2016)}]{goodfellow2016deep}
Goodfellow, I.; Bengio, Y.; and Courville, A. 2016.
\newblock \emph{Deep learning}.
\newblock MIT press.

\bibitem[{He et~al.(2017)He, Liao, Zhang, Nie, Hu, and Chua}]{he2017neural}
He, X.; Liao, L.; Zhang, H.; Nie, L.; Hu, X.; and Chua, T.-S. 2017.
\newblock Neural collaborative filtering.
\newblock In \emph{Proceedings of the 26th international conference on world
  wide web}, 173--182.

\bibitem[{Hinton(1999)}]{hinton1999products}
Hinton, G.~E. 1999.
\newblock Products of experts .

\bibitem[{Hu et~al.(2019)Hu, Yu, Tu, Yang, Chen, and Dai}]{hu2019multi}
Hu, Y.-Q.; Yu, Y.; Tu, W.-W.; Yang, Q.; Chen, Y.; and Dai, W. 2019.
\newblock Multi-Fidelity Automatic Hyper-Parameter Tuning via Transfer Series
  Expansion.
\newblock \emph{AAAI} .

\bibitem[{Hutter, Hoos, and Leyton-Brown(2011)}]{hutter2011sequential}
Hutter, F.; Hoos, H.~H.; and Leyton-Brown, K. 2011.
\newblock Sequential model-based optimization for general algorithm
  configuration.
\newblock In \emph{International Conference on Learning and Intelligent
  Optimization}, 507--523. Springer.

\bibitem[{Hutter, Kotthoff, and Vanschoren(2018)}]{automl_book}
Hutter, F.; Kotthoff, L.; and Vanschoren, J., eds. 2018.
\newblock \emph{Automated Machine Learning: Methods, Systems, Challenges}.
\newblock Springer.
\newblock In press, available at http://automl.org/book.

\bibitem[{Jamieson and Talwalkar(2016)}]{jamieson2016non}
Jamieson, K.; and Talwalkar, A. 2016.
\newblock Non-stochastic best arm identification and hyperparameter
  optimization.
\newblock In \emph{Artificial Intelligence and Statistics}, 240--248.

\bibitem[{Jiang et~al.(2017)Jiang, Jiang, Cui, and
  Zhang}]{jiang2017tencentboost}
Jiang, J.; Jiang, J.; Cui, B.; and Zhang, C. 2017.
\newblock TencentBoost: a gradient boosting tree system with parameter server.
\newblock In \emph{2017 IEEE 33rd ICDE}, 281--284. IEEE.

\bibitem[{Jones, Schonlau, and Welch(1998)}]{jones1998efficient}
Jones, D.~R.; Schonlau, M.; and Welch, W.~J. 1998.
\newblock Efficient global optimization of expensive black-box functions.
\newblock \emph{Journal of Global optimization} 13(4): 455--492.

\bibitem[{Kandasamy et~al.(2017)Kandasamy, Dasarathy, Schneider, and
  Poczos}]{kandasamy2017multi}
Kandasamy, K.; Dasarathy, G.; Schneider, J.; and Poczos, B. 2017.
\newblock Multi-fidelity bayesian optimisation with continuous approximations.
\newblock \emph{arXiv preprint arXiv:1703.06240} .

\bibitem[{Klein et~al.(2017{\natexlab{a}})Klein, Falkner, Bartels, Hennig, and
  Hutter}]{kleinfbhh17}
Klein, A.; Falkner, S.; Bartels, S.; Hennig, P.; and Hutter, F.
  2017{\natexlab{a}}.
\newblock Fast Bayesian Optimization of Machine Learning Hyperparameters on
  Large Datasets.
\newblock In \emph{Proceedings of the 20th International Conference on
  Artificial Intelligence and Statistics}, 528--536.

\bibitem[{Klein et~al.(2017{\natexlab{b}})Klein, Falkner, Springenberg, and
  Hutter}]{klein2016learning}
Klein, A.; Falkner, S.; Springenberg, J.~T.; and Hutter, F. 2017{\natexlab{b}}.
\newblock Learning curve prediction with Bayesian neural networks.
\newblock \emph{Proceedings of the International Conference on Learning
  Representations} .

\bibitem[{Li et~al.(2018)Li, Jamieson, DeSalvo, Rostamizadeh, and
  Talwalkar}]{li2018hyperband}
Li, L.; Jamieson, K.; DeSalvo, G.; Rostamizadeh, A.; and Talwalkar, A. 2018.
\newblock Hyperband: A novel bandit-based approach to hyperparameter
  optimization.
\newblock \emph{Proceedings of the International Conference on Learning
  Representations} 1--48.

\bibitem[{Li et~al.(2020)Li, Jiang, Gao, Shao, Zhang, and
  Cui}]{li2020efficient}
Li, Y.; Jiang, J.; Gao, J.; Shao, Y.; Zhang, C.; and Cui, B. 2020.
\newblock Efficient Automatic CASH via Rising Bandits.
\newblock In \emph{AAAI}, 4763--4771.

\bibitem[{Ma et~al.(2019)Ma, Wen, Zhong, Chen, and Li}]{ma2019mmm}
Ma, J.; Wen, J.; Zhong, M.; Chen, W.; and Li, X. 2019.
\newblock MMM: Multi-source Multi-net Micro-video Recommendation with Clustered
  Hidden Item Representation Learning.
\newblock \emph{Data Science and Engineering} 4(3): 240--253.

\bibitem[{Poloczek, Wang, and Frazier(2017)}]{poloczek2017multi}
Poloczek, M.; Wang, J.; and Frazier, P. 2017.
\newblock Multi-information source optimization.
\newblock In \emph{Advances in Neural Information Processing Systems},
  4288--4298.

\bibitem[{Schilling et~al.(2015)Schilling, Wistuba, Drumond, and
  Schmidt-Thieme}]{schilling2015hyperparameter}
Schilling, N.; Wistuba, M.; Drumond, L.; and Schmidt-Thieme, L. 2015.
\newblock Hyperparameter optimization with factorized multilayer perceptrons.
\newblock In \emph{Joint European Conference on Machine Learning and Knowledge
  Discovery in Databases}, 87--103. Springer.

\bibitem[{Schilling, Wistuba, and Schmidt-Thieme(2016)}]{schilling2016scalable}
Schilling, N.; Wistuba, M.; and Schmidt-Thieme, L. 2016.
\newblock Scalable hyperparameter optimization with products of gaussian
  process experts.
\newblock In \emph{Joint European conference on machine learning and knowledge
  discovery in databases}, 33--48. Springer.

\bibitem[{Sen, Kandasamy, and Shakkottai(2018)}]{sen2018noisy}
Sen, R.; Kandasamy, K.; and Shakkottai, S. 2018.
\newblock Noisy Blackbox Optimization with Multi-Fidelity Queries: A Tree
  Search Approach.
\newblock \emph{arXiv: Machine Learning} .

\bibitem[{Snoek, Larochelle, and Adams(2012)}]{snoek2012practical}
Snoek, J.; Larochelle, H.; and Adams, R.~P. 2012.
\newblock Practical bayesian optimization of machine learning algorithms.
\newblock In \emph{Advances in neural information processing systems}.

\bibitem[{Swersky, Snoek, and Adams(2013)}]{swersky2013multi}
Swersky, K.; Snoek, J.; and Adams, R.~P. 2013.
\newblock Multi-task bayesian optimization.
\newblock In \emph{Advances in neural information processing systems},
  2004--2012.

\bibitem[{Swersky, Snoek, and Adams(2014)}]{swersky2014freeze}
Swersky, K.; Snoek, J.; and Adams, R.~P. 2014.
\newblock Freeze-thaw Bayesian optimization.
\newblock \emph{arXiv preprint arXiv:1406.3896} .

\bibitem[{Takeno et~al.(2020)Takeno, Fukuoka, Tsukada, Koyama, Shiga, Takeuchi,
  and Karasuyama}]{takeno2020multifidelity}
Takeno, S.; Fukuoka, H.; Tsukada, Y.; Koyama, T.; Shiga, M.; Takeuchi, I.; and
  Karasuyama, M. 2020.
\newblock Multi-fidelity Bayesian Optimization with Max-value Entropy Search
  and its parallelization.

\bibitem[{Wang, Xu, and Wang(2018)}]{combine_bohb1}
Wang, J.; Xu, J.; and Wang, X. 2018.
\newblock Combination of Hyperband and Bayesian Optimization for Hyperparameter
  Optimization in Deep Learning .

\bibitem[{Wang et~al.(2013)Wang, Zoghi, Hutter, Matheson, and
  De~Freitas}]{wang2013bayesian}
Wang, Z.; Zoghi, M.; Hutter, F.; Matheson, D.; and De~Freitas, N. 2013.
\newblock Bayesian optimization in high dimensions via random embeddings.
\newblock In \emph{Twenty-Third International Joint Conference on Artificial
  Intelligence}.

\bibitem[{Wistuba, Schilling, and Schmidt-Thieme(2016)}]{wistuba2016two}
Wistuba, M.; Schilling, N.; and Schmidt-Thieme, L. 2016.
\newblock Two-stage transfer surrogate model for automatic hyperparameter
  optimization.
\newblock In \emph{Joint European conference on machine learning and knowledge
  discovery in databases}, 199--214. Springer.

\bibitem[{Wu et~al.(2019{\natexlab{a}})Wu, Toscano-Palmerin, Frazier, and
  Wilson}]{Wu2019}
Wu, J.; Toscano-Palmerin, S.; Frazier, P.~I.; and Wilson, A.~G.
  2019{\natexlab{a}}.
\newblock {Practical multi-fidelity Bayesian optimization for hyperparameter
  tuning}.

\bibitem[{Wu et~al.(2019{\natexlab{b}})Wu, Toscanopalmerin, Frazier, and
  Wilson}]{wu2019practical}
Wu, J.; Toscanopalmerin, S.; Frazier, P.~I.; and Wilson, A.~G.
  2019{\natexlab{b}}.
\newblock Practical multi-fidelity Bayesian optimization for hyperparameter
  tuning 284.

\bibitem[{Wu et~al.(2020)Wu, Zhang, Gao, Bian, and Cui}]{Wu2020}
Wu, S.; Zhang, Y.; Gao, C.; Bian, K.; and Cui, B. 2020.
\newblock {GARG: Anonymous Recommendation of Point-of-Interest in Mobile
  Networks by Graph Convolution Network}.
\newblock \emph{Data Science and Engineering} ISSN 23641541.
\newblock \doi{10.1007/s41019-020-00135-z}.

\bibitem[{Yao et~al.(2018)Yao, Wang, Escalante, Guyon, Hu, Li, Tu, Yang, and
  Yu}]{automl}
Yao, Q.; Wang, M.; Escalante, H.~J.; Guyon, I.; Hu, Y.; Li, Y.; Tu, W.; Yang,
  Q.; and Yu, Y. 2018.
\newblock Taking Human out of Learning Applications: {A} Survey on Automated
  Machine Learning.
\newblock \emph{CoRR} .

\bibitem[{Yogatama and Mann(2014)}]{yogatama2014efficient}
Yogatama, D.; and Mann, G. 2014.
\newblock Efficient transfer learning method for automatic hyperparameter
  tuning.
\newblock In \emph{Artificial Intelligence and Statistics}, 1077--1085.

\bibitem[{Zhang et~al.(2020)Zhang, Jiang, Shao, and
  Cui}]{DBLP:journals/chinaf/ZhangJSC20}
Zhang, W.; Jiang, J.; Shao, Y.; and Cui, B. 2020.
\newblock Snapshot boosting: a fast ensemble framework for deep neural
  networks.
\newblock \emph{Sci. China Inf. Sci.} 63(1): 112102.
\newblock \doi{10.1007/s11432-018-9944-x}.
\newblock \urlprefix\url{https://doi.org/10.1007/s11432-018-9944-x}.

\bibitem[{Z{\"{o}}ller and Huber(2019)}]{DBLP:journals/corr/abs-1904-12054}
Z{\"{o}}ller, M.; and Huber, M.~F. 2019.
\newblock Survey on Automated Machine Learning.
\newblock \emph{CoRR} abs/1904.12054.

\end{thebibliography}

\end{document}